# The Semantic Ladder: A Framework for Progressive Formalization of Natural Language Content for Knowledge Graphs and AI Systems

Vogt, Lars[1]  orcid.org/0000-0002-8280-0487

[1]*Leibniz Institute for the Analysis of Biodiversity Change (LIB), Museum of Nature Hamburg, Martin-8 Luther-King Platz 3, 20146 Hamburg, Germany*

# Abstract

Semantic data and knowledge infrastructures must reconcile two fundamentally different forms of representation: natural language, in which most knowledge is created and communicated, and formal semantic models, which enable machine-actionable integration, interoperability, and reasoning. Bridging this gap remains a central challenge, particularly when full semantic formalization is required at the point of data entry.

Here, we introduce the Semantic Ladder, an architectural framework that enables the progressive formalization of data and knowledge. Building on the concept of modular semantic units as identifiable carriers of meaning, the framework organizes representations across levels of increasing semantic explicitness, ranging from natural language text snippets to ontology-based and higher-order logical models. Transformations between levels support semantic enrichment, statement structuring, and logical modelling while preserving semantic continuity and traceability.

This approach enables the incremental construction of semantic knowledge spaces, reduces the semantic parsing burden, and supports the integration of heterogeneous representations, including natural language, structured semantic models, and vector-based embeddings. The Semantic Ladder thereby provides a foundation for scalable, interoperable, and AI-ready data and knowledge infrastructures.

# 1. Introduction

Knowledge graphs have emerged as a foundational technology for representing, integrating, and reasoning over heterogeneous data and knowledge across domains (1). Their explicit semantics and graph-based structure make them well-suited for semantic interoperability, knowledge discovery, and advanced querying, and they are increasingly used in scientific, enterprise, and public knowledge infrastructures such as [Wikidata](). In parallel, the **FAIR Principles** (**F**indable, **A**ccessible, **I**nteroperable, **R**eusable) (2) have become a central framework for guiding the publication and management of scientific data. Knowledge graphs and ontologies are widely regarded as key enablers of FAIRness due to their support for rich metadata, semantic linking, SPARQL querying, and standardized representations (3–5).

Despite these advances, constructing and revising large-scale knowledge graphs remains challenging in practice. Two persistent barriers limit their scalability and broader adoption. First, the **knowledge acquisition bottleneck and semantic parsing burden** (6,7) arises from the need to transform data and knowledge expressed in natural language into structured, semantically grounded representations using the *Subject—Predicate—Object* triple syntax of the Resource Description Framework ([RDF]()) and the logical formalism of the Web Ontology Language ([OWL]()), which is based on Description Logics. This process typically requires substantial modelling expertise and close collaboration between domain experts and ontology engineers, making it time-consuming and difficult to scale. Second, to be semantically interoperable, most knowledge graph approaches assume **predefined and relatively static schemata**, requiring upfront formal modelling before data and knowledge can be integrated. This assumption conflicts with the iterative and evolving nature of scientific discovery and hinders **dynamic knowledge graph construction**, where data and knowledge are incrementally added and refined over time (7).

As a consequence, participation in knowledge graph creation is often restricted to technically skilled contributors, growth remains slow, and **premature formalization** occurs, i.e., requiring ontology and modelling commitments before the underlying semantic content is fully articulated, and its practical usage is known. These limitations constrain both the scalability and the openness of OWL-based knowledge infrastructures.

Moreover, not all data and knowledge require the level of semantic formalization needed for logical reasoning. In many scenarios, the primary objective is to make information findable, accessible, and human-interpretable, rather than to enable full logical inference and machine-interpretability. Requiring complete ontology-based modelling in such cases introduces unnecessary complexity and raises barriers for contributors, particularly in open, community-driven environments such as citizen science or interdisciplinary research.

To address these challenges, data and knowledge representations must move beyond **atomistic structural primitives** such as triples or table cells and instead be centred on **units of meaning** that reflect how humans naturally communicate and interpret data and knowledge. We refer to such representations as **semantic units**, defined as minimal, self-contained units of meaning that can be interpreted independently within a knowledge space (8). Aligning technical representations with these units reduces cognitive and technical barriers while preserving semantic clarity and interoperability. This perspective is consistent with the **CLEAR Principle** (**C**ognitively interoperable, semantically **L**inked, contextually **E**xplorable, intuitively **A**ccessible, and human-**R**eadable and -interpretable) (9) and with emerging extensions of the FAIR Principles that emphasize propositional interoperability (3).

Building on this foundation, we propose a framework for **progressive semantic formalization** in data and knowledge infrastructures. Rather than requiring full formal modelling at the point of entry, the framework allows data and knowledge to be captured at different levels of semantic formalization and to evolve toward more formal representations as needed. This approach reflects the way scientific knowledge is typically produced and communicated: beginning with informal descriptions, progressing through structured statements, and ultimately becoming formalized within logical and ontological frameworks. It also enables human-AI collaboration, in which automated methods such as Large Language Models (LLMs) can assist in enriching and formalizing data and knowledge while domain experts remain the arbiters of meaning.

This paper proposes a **conceptual architectural framework** that builds on these principles and makes four contributions:

1. **Progressive semantic formalization:** We introduce the **Semantic Ladder** as a principled mechanism that allows data and knowledge to enter a system at different levels of semantic formalization and to be incrementally transformed toward more semantically formalized representations.
2. **Meaning-centred, technology-agnostic knowledge architecture:** We extend the technology-agnostic **Semantic Units Framework** (7,8,10) by introducing text snippets and Rosetta Statements as first-class semantic units, establishing formalized natural language statements as semantic anchors that bridge natural language expressions and logic-based representations.
3. **Multi-representation semantic equivalence:** The framework supports coexistence of semantically equivalent representations across different levels of formalization, preserving semantic continuity, provenance, and semantic interpretability across heterogeneous modelling approaches.
4. **Integration of symbolic and vector knowledge spaces:** We enable hybrid data and knowledge infrastructures in which semantic units simultaneously support symbolic reasoning and embedding-based AI methods, allowing structured knowledge graphs and modern AI systems to operate within a unified semantic space.

Together, these contributions define a **unified, meaning-centred, technology-agnostic architecture** in which data and knowledge can be represented, enriched, and interpreted across multiple levels of semantic commitment. This architecture enables a unified semantic knowledge space in which empirical observations, formal definitions, hypotheses, and research questions can coexist as semantic units. By supporting incremental enrichment and representation-level flexibility, the framework facilitates **dynamic knowledge graph construction**, reduces the semantic parsing burden, and improves FAIRness, CLEARness, and AI-readiness.

The remainder of this paper elaborates the conceptual foundations of the Semantic Units Framework, introduces the Semantic Ladder in detail, and discusses its implications for dynamic knowledge graph construction, hybrid AI-knowledge systems, and FAIR- and CLEAR-aligned data and knowledge infrastructures.

This work is conceptual in nature and focuses on the design principles and architectural foundations of semantic data and knowledge infrastructures rather than on a specific system implementation. By abstracting from particular technologies, the proposed framework is intended to provide a generalizable foundation that can guide the development of future knowledge systems and

AI-integrated data infrastructures. The framework is designed to be compatible with existing knowledge graph technologies, ontology languages, and AI-based methods, enabling incremental adoption in real-world systems.

## 2. Background and Conceptual Context

### 2.1 Knowledge Graph Construction and Semantic Modelling

Knowledge graphs have traditionally been constructed through ontology-driven semantic modelling approaches that rely on formally defined schemata and controlled vocabularies specified in OWL. In such approaches, ontologies define classes, properties, and logical constraints that govern the structure of the knowledge graph and enable automated reasoning. This paradigm has enabled the creation of large collaborative knowledge bases such as Wikidata, as well as domain-specific knowledge infrastructures.

In most knowledge graph systems, the basic structural unit is the RDF triple, which represents data and knowledge as a binary *Subject—Predicate—Object* relation. Ontology-based construction typically follows a schema-first methodology. While this provides strong formal semantics, it introduces substantial modelling overhead. To mitigate this, practitioners have increasingly turned to **constraint-based validation frameworks** such as the **Shapes Constraint Language (SHACL)** (11) and modelling languages like **LinkML** (12,13). These frameworks allow for a more pragmatic, "closed-world" validation of data shapes without the full computational complexity of OWL reasoning. However, even with these tools, the **semantic parsing burden** remains, as the technical shapes first must be developed for each new type of information, before this information can be added to the graph, and domain experts must still map their data and knowledge to these shapes. Both requires substantial knowledge in formal semantics and experience with the existing ontology landscape. Integrating new types of information often requires prior modelling decisions that may become obsolete as the domain evolves, leading to the "premature formalization" mentioned previously.

### 2.2 FAIR Knowledge Infrastructures

The FAIR Principles (2) emphasize the use of persistent identifiers, standardized metadata, and shared vocabularies. Knowledge graphs and ontologies are recognized as the primary technical vehicle for FAIRness because they facilitate the integration of heterogeneous datasets through shared identifiers (5). Large-scale initiatives, including collaborative resources such as [Wikidata](#) (14) and [MaterialDigital](#) (15), as well as various FAIR Digital Object (FDO) implementations (16,17), have adopted graph-based representations to bridge the gap between raw data and machine-actionable knowledge.

However, implementing FAIR at scale requires more than structured metadata. Recent work emphasizes **propositional interoperability**, i.e., the ability for machines to not only find and interpret individual ontology terms and binary relations within a dataset but to find and interpret the specific claims expressed within it (3). Other work emphasizes **granular FAIRness**, i.e., that representing entities with a high granular complexity results in complex and granularly nested representations

(18). This highlights the need for representation models to capture semantically coherent units of meaning that can be organized into nested modular structures rather than isolated data points.

## 2.3 Natural Language and Knowledge Representation

Because much human information is expressed in natural language, numerous approaches extract structured representations from natural language text. Traditional information extraction and relation extraction have recently been augmented by **Large Language Models (LLMs)**, which identify entities and candidate triples with high linguistic fluency. LLMs are increasingly used to support the construction of knowledge graphs and ontologies, and they have been integrated with knowledge graphs to improve their own reasoning capabilities to reduce their hallucinations (19–26).

Despite these gains, a "semantic gap" persists. Automatically extracted knowledge often requires manual validation and alignment to ensure it adheres to the logical consistency required by high-quality, semantically precise ontologies and knowledge graphs suitable for scientific use. Ambiguity and context-dependency mean that a triple extracted by an LLM may lack the necessary provenance or epistemic context (e.g., distinguishing a "working assumption" from a "proven fact") required for scientific rigour The challenge remains to bridge these "noisy" natural language observations with formally structured semantic representations while ensuring semantic precision and consistency (27,28).

## 2.4 Statement-Based Knowledge Representation

Human knowledge communication, particularly in scientific discourse, is largely expressed through statements describing observations, claims, procedures, definitions, or hypotheses. Statements are the primary meaning-carrying units in human symbolic communication. Treating statements as first-class semantic objects allows knowledge structures to preserve the meaning of propositions, support provenance, and enable modular, semantically coherent organization of knowledge.

Several approaches treat statements as fundamental units. For example, Nanopublications (17,29–31) represent assertions together with their provenance and publication metadata as small, citable RDF graphs using Named Graphs (32). These approaches demonstrate how statement-level representations can support FAIR and machine-actionable data and knowledge publication.

Building on these ideas, the **Semantic Units Framework** organizes data and knowledge into semantically coherent, self-contained units of meaning (8,10). This framework is described in detail in the following chapter.

**Rosetta Statements** (7) provide a structured approach for representing natural language statements as formalized patterns in RDF, bridging natural language expressions and formal semantic models.

Traditional ontology architectures distinguish between the conceptual knowledge (TBox) and instance-level assertions (ABox). Alternative approaches seek to integrate these forms of knowledge more tightly within **unified representation frameworks** (10).

Supporting such heterogeneous forms of data and knowledge requires a representation architecture that enables progressive semantic enrichment while maintaining coherence across representations. The following chapter introduces the Semantic Units Framework as a foundation for such an architecture.

# 3. The Semantic Units Framework

The Semantic Ladder introduced in this paper builds upon the **Semantic Units Framework** developed in earlier work (8,10,18). The framework conceptualizes data and knowledge as collections of modular, identifiable units of meaning that serve as the fundamental building blocks of information infrastructures. Here, we summarize its core principles and introduce extensions required for the Semantic Ladder, in particular the incorporation of text snippet and Rosetta Statement semantic units as explicit representation forms.

Existing data infrastructures operate on **structural primitives** as basic building blocks of representation. RDF-based systems operate on triples, property graphs on nodes and edges, and relational databases on rows and cells in tables. These primitives primarily reflect data organization rather than the meaning of the content. The Semantic Units Framework addresses this limitation by focusing on **units of meaning** by following the **semantic modularization principle**, in which data and knowledge are partitioned into independently addressable, semantically coherent units. Each such unit (i) may apply its own logical framework or no framework at all for modelling its semantic content, enabling distributed reasoning across heterogeneous components of a knowledge infrastructure (**logical heterogeneity across modules**), (ii) is self-contained, with its semantic content being internally consistent and meaningful to a human (as opposed to individual triples, which are not always meaningful; cf. triple highlighted in red in Fig. 2) (**semantic coherence within a module**), (iii) is uniquely identifiable and can be reused, cited, and queried independently (**independent addressability**), (iv) is human-interpretable with sets of units being navigable and potentially interlinked to form larger coherent units that support contextual exploration (**cognitive interoperability and contextual explorability**), and (v) can reference other units, resulting in recursively nested units that preserve modular semantics and granular FAIRness (**reusability and composability**) (10,18).

Semantic modularization thereby reflects how humans manage conceptual complexity. In scientific discourse, complex phenomena are typically encapsulated through terms that act as semantic handles (e.g., "Darwinian evolution", "COVID-19", or "information bubble"), enabling them to function as identifiable entities in reasoning and communication. The Semantic Units Framework translates this strategy into data and knowledge infrastructure design (18), with semantic units functioning as the **fundamental representational building blocks** of the framework, enabling data and knowledge to be organized as discrete, reusable, and composable units of meaning.

## 3.1 Semantic Units as Modular Units of Meaning

A **semantic unit** is a bounded, identifiable structure that carries a coherent unit of meaning. Semantic units may represent individual propositions or higher-level knowledge structures that organize multiple semantic units into meaningful aggregates (8).

More formally, let $K$ denote a semantic data and knowledge space and $E$ the set of entities referenced within it, including all individuals, classes (i.e., concepts, universals, types), properties (i.e., relations), and semantic units. A semantic unit can be defined as the tuple

$$SU = (t, c, R, D, M).$$

Here, $t$ denotes the unit type, $c$ its content representation (=meaning), $R \subseteq E$ the entities referenced in this $SU$, $D$ the modelling schema for $c$, and $M$ the associated metadata, including contextual and provenance information. A semantic unit therefore represents a minimal independently interpretable unit of meaning within the knowledge space $K$. $D$ can thereby range from natural language, a schema for relational databases based on SQL Data Definition Language (SQL DDL) (33,34), a graph schema for RDF/OWL based on SHACL (11) or the Linked Data Modeling Language (LinkML) (12,13), or a JSON Schema for JSON-based systems.

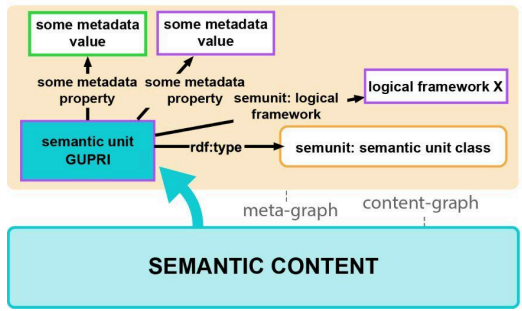

**Figure 1: Core structure of a semantic unit, exemplified within an OWL/RDF framework and a tabular/relational framework. (a)** A graph-based implementation of a semantic unit, with the unit's semantic content (meaning-carrying triples not shown) being stored in a Named Graph (32), constituting the unit's content-graph. The Named Graph's GUPRI is also the GUPRI of the semantic unit (indicated by the blue arrow). All contextual information relating to the semantic unit itself, including the class it instantiates, its provenance, and other metadata, are stored in the unit's meta-graph. **(b)** A tabular- or relational database-based implementation of a semantically equivalent semantic unit. The information from the content-graph of a) is organized in the content table, and the information from the meta-graph in the meta table. *Figure adapted from (10).*

In practice, a semantic unit corresponds to a structured subset of data or knowledge explicitly assigned a **Globally Unique Persistent and Resolvable Identifier (GUPRI)**, i.e., the semantic unit resource (Fig. 1). This resource enables the unit to function as a referenceable and reusable digital object. Every semantic unit resource is thereby an instance of a corresponding semantic unit class. For example, every semantic unit representing a measurement would be an instance of a *measurement* class (8,10).

Semantic content of a unit is separated from its contextual metadata. In graph-based systems, this corresponds to **content-graphs** and **meta-graphs**; in relational database systems, to content and meta tables (cf. Fig. 1).

This separation between semantic content and contextual metadata enables knowledge infrastructures to maintain clear semantic boundaries while supporting rich annotation and provenance tracking. At the same time, the explicit identity of semantic units allows them to function as modular components that can be referenced and combined to form more complex knowledge structures.

## 3.2 Types of Semantic Units

The Semantic Units Framework distinguishes between two primary categories of semantic units: statement units and compound units (8,10).

**Statement units** represent individual propositions as semantically coherent objects. They correspond to minimal, complete statements that group all structural elements required to express a specific complete claim, observation, or relationship.

Individual triples, database cells, or graph edges may carry partial pieces of information, but they typically lack meaning in isolation (see triple highlighted in red, Fig. 2). Statement units therefore provide a mechanism for representing complete propositions as reusable semantic artefacts within a knowledge infrastructure.

### a) Natural Language Statement

*Specimen X has a mass of 4.96 grams*

### b) Graph Statement Unit

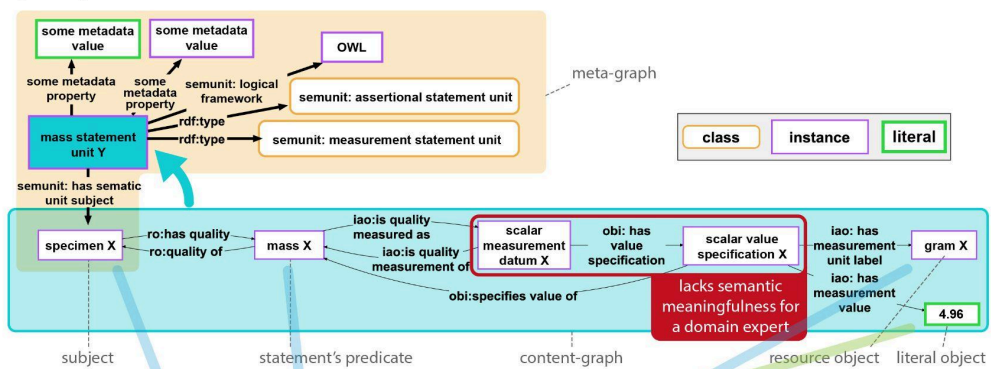

### c) Tabular Statement Unit

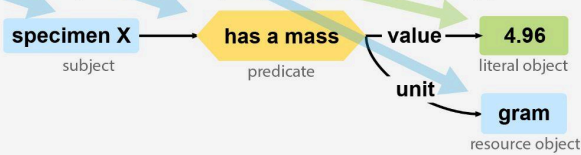

**Figure 2: From natural language statement to graph and tabular statement unit. (a)** A human-readable statement about the mass measurement of specimen X. **(b)** The representation of the same statement as an OWL-based statement unit, with the semantic content organized within the content-graph (light-blue box with blue border), adhering to RDF syntax and following the established pattern for measurement data from the Ontology of Biomedical Investigations (OBI) (35). The content-graph contains the statement with '*specimen X*' as its subject and '*mass X*' and '*gram X*' alongside the numerical value of 4.96 as its objects. The peach-coloured box depicts the unit's meta-graph, explicitly denoting the statement unit resource (blue box with purple border) as an instance of the classes *SEMUNIT:measurement statement unit* and *SEMUNIT:assertional statement unit*, with '*specimen X*' identified as the statement's subject. Note that the GUPRI of the statement unit is also the GUPRI of the semantic unit's content-graph (indicated by the opaque blue arrow). The meta-graph contains various metadata triples, here only placeholders shown. Highlighted in red within the content-graph is an example of a triple that is required for modelling purposes but lacks semantic meaningfulness for a human user. **(c)** The semantic content from a) and b), modelled as a tabular statement unit consisting of two tables, i.e., a content and a meta table. In a relational database, a statement unit is documented by a single row in each of these tables. **(d)** The dynamic label and **(e)** dynamic graph associated with the statement units from b) and c). The transparent blue and green arrows

indicate the mappings between slots in the graph schema, cells in the table, and slots in the dynamic label and dynamic graph templates, indicating semantic equivalence across the respective slots. *Figure adapted from* (10).

**Compound units** organize collections of statement units into higher-level semantic structures, such as object or process descriptions, arguments, or datasets (Fig. 3). By referencing sets of semantic units, compound units provide a mechanism for representing larger semantic structures while preserving the modular integrity of the underlying statement units.

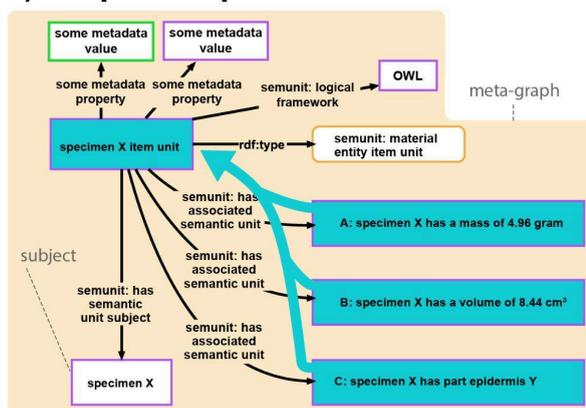

Figure 3: Example of a compound unit, denoted as '*specimen X item unit*'. (a) The RDF-based compound unit. (b) A semantically equivalent implementation as tabular compound unit. Both versions encompass the same set of statement units (A-C) describing specimen X. By virtue of merging the content-graphs or content tables of their associated statement units, compound units only indirectly possess a content-graph or content table. Consequently, compound units only specify meta-graphs (depicted in the peach-coloured box) or meta tables that reference the associated semantic units. *Figure adapted from* (10).

Together, statement units and compound units allow knowledge infrastructures to represent semantic content at multiple levels of granularity while preserving their semantic integrity, therewith avoiding the problem of meaning fragmentation that is otherwise inherent in the RDF triple syntax (10). Individual propositions can be represented as atomic statement units, while larger knowledge structures emerge through the composition and organization of these units into compound units.

## 3.3 Technology-Agnostic Representation

The Semantic Units Framework defines an abstraction representation layer that is **independent of specific technologies**. Semantic units can be implemented across RDF/OWL knowledge graphs, labelled property graphs, or relational databases while retaining their role as identifiable containers of meaning (10).

Different implementation technologies realize this abstraction in different ways. In RDF-based environments, a semantic unit may be represented as a **Named Graph** (32), whose identifier corresponds to the units' GUPRI. In relational database implementations, the semantic content and associated metadata of a unit may be distributed across related tables that jointly represent the unit's structure (cf. Fig. 1-3). Other implementations may use JSON-based objects. Although the technical realizations differ, the conceptual separation between **semantic content** and **contextual metadata** remains constant across implementations, with each semantic unit possessing its own identifier (GUPRI) that instantiates a corresponding semantic unit class.

This abstraction layer ensures that the framework remains robust against changes in underlying technologies. As data infrastructures evolve, the conceptual organization of knowledge into semantic units can remain stable, preserving semantic interoperability across heterogeneous systems.

While the Semantic Units Framework is independent of specific implementation technologies, the **process of translating natural language knowledge into structured representations** reveals important differences between alternative representation forms. Figure 4 illustrates this relationship for the example of a measurement statement.

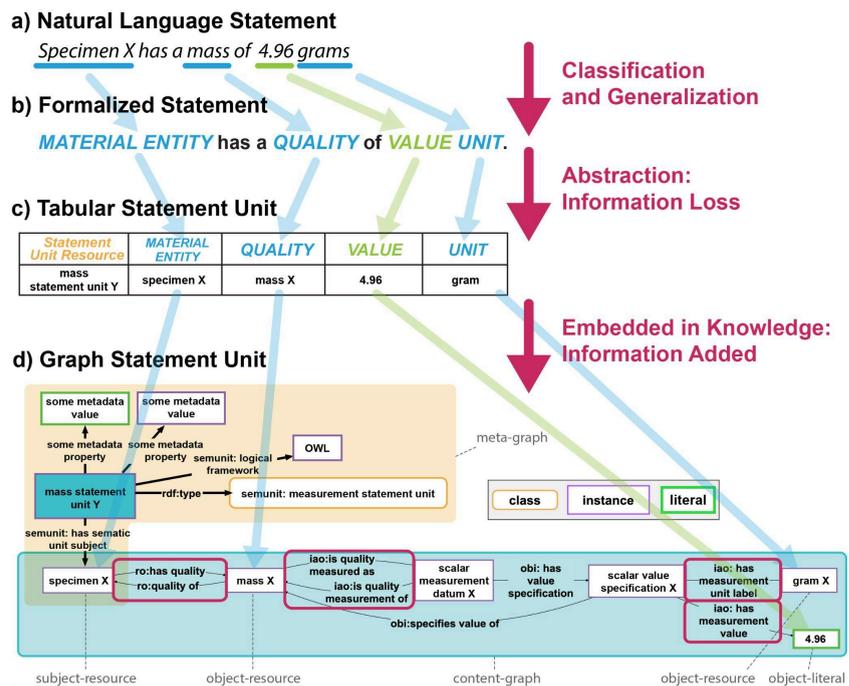

**Figure 4: Relation between natural language statements, formalized statements, tabular statement units, and RDF/OWL-based statement units.** **(a)** A natural language measurement statement with its syntactic positions highlighted. **(b)** Formalized statement obtained by classifying and generalizing the terms occupying each syntactic position. This formalized statement functions as a metamodel describing the general structure of measurement statements with the highlighted terms describing the position's semantic roles. **(c)** Tabular statement unit derived from the formalized statement. Each column corresponds to a semantic role/position of the formalized statement, and the ontology classes that define the semantic roles specifying the column constraint. The explicit relationships between positions are not represented in the table structure. **(d)** RDF/OWL-based statement unit representation of the same statement, in which the semantic relationships between the statement positions are explicitly encoded using ontology relations (marked red).

A natural language statement such as '*Specimen X has a mass of 4.96 grams*' can be decomposed into semantic positions with associated roles (Fig. 4a). By **classifying and generalizing** these roles using ontology concepts, a **formalized statement pattern** can be derived. For example, the subject '*specimen X*' may be classified as an instance of '*specimen*' from the Ontology for Biomedical Investigations (OBI:0100051) and further generalized along the OBI taxonomy to 'material entity' from the Basic Formal Ontology (BFO:0000040). Applying this to all statement positions results in the formalized statement '*MATERIAL ENTITY has a QUALITY of VALUE UNIT*' (Fig. 4b).

This pattern can be represented in **tabular form**, where positions correspond to columns. The ontology class defining the semantic role functions in the table as the constraint for the corresponding column (Fig. 4c). The semantic structure of the natural language statement is thereby preserved only implicitly, through a shared interpretation of the table schema. After translating a natural language statement into a tabular representation, **information about the semantic**

**relationships between the statement's positions is no longer explicitly represented in the table**, as the table only captures the positions of the statement but not the relationships that connect them.

In contrast, the semantically equivalent **RDF/OWL-based representation** of the statement preserves these relationships, as the entities representing the subjects and objects of the statement are connected through formally defined ontology relations (Fig. 4d). Consequently, tabular statement units are only **semantically equivalent** to their OWL-based counterparts if the relationships between the statement positions are known.

This illustrates a central aspect of the **semantic parsing burden**: when semantic relations are not explicitly represented in the source data, they must be reconstructed during the knowledge graph construction process. By organizing knowledge around semantic units and formalized statement patterns, the Semantic Units Framework provides a conceptual structure that helps reduce this burden by making the underlying semantic roles of statement components explicit.

## 3.4 Human-Interpretable Representations of Semantic Units

Semantic units support both machine- and human-interpretable representations. **Dynamic textual labels** render the semantic content of semantic units as natural language statements (Fig. 2d), while **dynamic graph patterns** specify how the content should be visualized graphically in user interfaces (Fig. 2e). Both display only the information that is meaningful for human interpretation and suppress structural elements that are primarily required for machine-actionable representations (cf. Fig. 2).

This separation between **machine-oriented representation** and **human-oriented display** of semantic content ensures that semantic units remain interpretable across different levels of formalization and supports both FAIR and CLEAR Principles.

## 3.5 Representation-Level Independence of Semantic Units

We extend the Semantic Units Framework by introducing **representation-level independence**: semantic units are independent not only of implementation technologies but also of the logical formalism used to represent their content. The same semantic content can be expressed using representations with different degrees of formalization while still being treated as semantic units within a unified knowledge space.

In practice, this means that a given proposition may be represented using several alternative modelling approaches that differ in their level of semantic formalization. At the lowest level, content may be captured as **text snippet semantic units**, where meaning is modelled as a natural language fragment extracted from a source document (Fig. 5a,b). Such units preserve the original linguistic formulation of knowledge while allowing it to be referenced, annotated, and contextualized within the knowledge space.

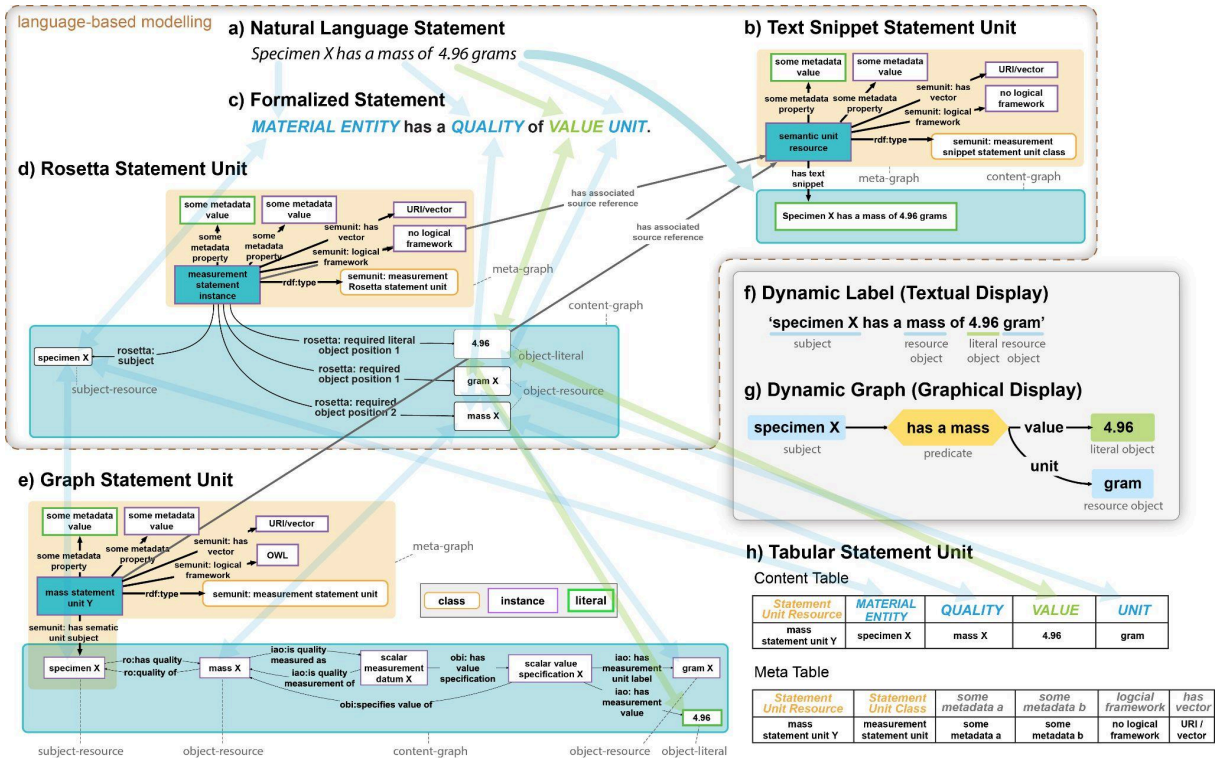

**Figure 5: Representation-level independence of the Semantic Units Framework.** (a) A natural language statement of a mass measurement of a particular specimen X. (b) The same statement represented as a text snippet statement unit. The unit's content-graph (light-blue box with blue border) carries the semantic content modelled as a natural language string, linked to the unit resource (blue box with purple border) via the data property '*has text snippet*'. The text snippet is vectorized, and the vector is either directly linked to the unit's resource or indirectly via an identifier, along with all embedding-related metadata (here not shown). Additional metadata include the reference to the source text, if applicable, along with start- and end-offset information to track the original location of the text in the publication. The metadata also indicates that no logical framework was applied for modelling the unit's content. (c) The formalized language schema that is derived from statement a), in which the subject and object positions have been classified and generalized (e.g., '*specimen X*' to '*MATERIAL ENTITY*' and '*mass*' to '*QUALITY*'). (d) Statement a), modelled as an RDF-based statement unit using the Rosetta Statement approach for modelling semantic content. Rosetta Statement units model the grammatical and syntactical structure of natural language sentences by applying the RDF reification method to the underlying sentence that is potentially n-ary (for Rosetta Statements, see (7)). Representations a)-d) have in common that they model the semantic content based on the structure of natural language: a)-c) use natural language directly and d) models natural language in RDF. (e) The same content as in a) and b), modelled in an OWL-based statement unit. Its content-graph is an OWL graph, with '*specimen X*' as its subject and '*mass X*', '*gram X*', and the numerical value '*4.96*' as its objects. Contrary to d), the graph aims to be a truthful representation of its real-world referent. d) and e) can be linked to a set of Text Snippet Statement units via the property '*has associated source reference*'. (f) The dynamic label and (g) dynamic graph associated with the measurement statement units d), e), and h) for representing their semantic contents in user interfaces. (h) The same statement modelled as a tabular statement unit. The blue and green arrows indicate the alignment of resource and literal slots across the different models, indicating their semantic equivalence, with the Rosetta Statement schema functioning as a reference schema and thus semantic anchor across all representations. *Figure adapted from* (18).

A more structured representation can be obtained through the structure provided by **formalized natural language statements**, which describe recurring patterns of information by identifying the semantic roles within a proposition (Fig. 5c; see also section 3.3). These patterns are represented as canonical statement structures that define the syntactic positions and semantic roles occurring in a particular **type of statement** (7). For example, a measurement statement may be formalized as a structure such as '*MATERIAL ENTITY has a QUALITY of VALUE UNIT*' (Fig. 5c), where each syntactic position defines a specific role within the statement.

Building on this concept, **Rosetta Statements schemata** provide canonical representations of formalized statements by decomposing a proposition into explicit positions with associated semantic roles (Fig. 5d). Rosetta Statement schemata were originally introduced as RDF-based natural language statement schemata serving as semantic anchors for aligning different OWL-based semantic schemata representing the same type of information, thereby reducing mapping complexity from *$n^2$* to *$2n$* (7). In the present work, we extend this concept by treating **Rosetta Statements as first-class semantic units** within the Semantic Units Framework. Consequently, their schemata not only function as schema-level mediation structures but also as **semantic anchors** that play a crucial **bridging role** between semantic units that represent their content in natural language and those that represent it using formal semantics and logical frameworks.

At higher levels of formalization, the same proposition may be represented using **logic-based semantic units**, such as OWL-based semantic units or other formal logical representations (Fig. 5e). In these representations, the semantic components of the statement are connected through formally defined ontology relations specified in the corresponding graph schema, allowing the knowledge graph to support automated reasoning, consistency checking, and logical inference.

By linking alternative representations through shared identifiers and explicit mappings, semantic units representing the same proposition can then coexist at different levels of formalization while remaining semantically anchored in a common statement structure represented as a Rosetta Statement unit. In this perspective, semantic units function as **representation-invariant carriers of meaning**, allowing the same proposition to be expressed across heterogeneous modelling paradigms. Although these alternative representations differ in their level of formalization, they represent **semantically equivalent units of meaning**. For human readers, they convey essentially the same proposition. For computational systems, however, the more formal representations provide additional semantic structure that supports interoperability, reasoning, and integration with ontological knowledge. This representation-level flexibility forms the conceptual foundation of the **Semantic Ladder** introduced in chapter 4.

## 3.6 Unified Semantic Knowledge Space

The modular architecture of semantic units also enables the construction of **unified semantic information spaces** capable of integrating multiple forms of data and knowledge representations within a single conceptual framework.

Traditional ontology-based knowledge systems distinguish between the **TBox** (terminological knowledge), which contains class axioms and universal statements, and instance-level assertional statements encoded in the **ABox**. While useful for formal logical reasoning, this separation limits the representation of the broader spectrum of statements found in scientific communication, including not only universal, assertional, existential, prototypical, directive, conditional, and interrogative statements, but also non-asserted contents, negations, absences, cardinality restrictions, disagreement, and epistemic beliefs (10). Within the Semantic Units Framework, these distinctions are not enforced through separate architectural layers but can instead be represented through **different types of semantic units** within a shared data and knowledge space. They can thus be represented within the same universe of discourse while preserving modularity, traceability, and semantic coherence.

Taken together, the concepts introduced in this chapter reveal that the Semantic Units Framework organizes data and knowledge infrastructures along **three orthogonal architectural dimension**.

**Semantic modularization** organizes data and knowledge as collections of semantic units, allowing content to be referenced, reused, and combined while preserving semantic identity. With their nested structure and their focus on statements as first-class semantic objects they prevent fragmentation of meaning, loss of contextual coherence, and support the incremental growth of knowledge graphs through **dynamic knowledge graph construction** (7).

**Technical independence** allows that semantic units can be implemented across heterogeneous data infrastructures, supporting **interoperability between different technical environments** (10). The conceptual organization of data and knowledge into semantic units therefore remains stable even when the underlying technical implementations vary.

**Representation-level independence** enables the same semantic content to be represented using different modelling approaches ranging from natural language expressions to formal logical models. This dimension forms the basis of the Semantic Ladder introduced in the following chapter and describes how knowledge can evolve from less formal to more formally structured representations while preserving semantic continuity, therewith lowering the entry-barrier for adding content to the knowledge space and thus addressing the **semantic parsing burden challenge** (7).

Together, these dimensions define the conceptual architecture of the Semantic Units Framework, enabling modular, technology-independent, and progressively formalized representations of data and knowledge. This foundation support the **Semantic Ladder**, which operationalizes progressive semantic formalization across representation levels.

# 4. The Semantic Ladder

## 4.1 Motivation: Progressive Semantic Formalization

While the previous chapter established that semantic units can be represented across multiple forms with different degrees of formalization, it does not specify how these representations are systematically organized or how semantic content transitions between them. In practice, data and knowledge rarely emerge in fully formalized form. Instead, they typically originate as natural language descriptions and are only progressively structured and formalized as modelling needs evolve. Requiring complete formalization at the point of data entry creates a significant barrier to participation and contributes to the semantic parsing burden discussed earlier.

To address this challenge, we introduce the **Semantic Ladder**, an architectural framework that organizes semantic units into an ordered progression of representation levels defined by increasing semantic explicitness and machine-interpretability. Rather than treating natural language and formal semantic models as separate or loosely connected layers, the Semantic Ladder integrates them within a unified structure that explicitly supports the gradual enrichment of semantic content.

Formally, the Semantic Ladder defines (i) a sequence of representation levels, (ii) transformation mechanisms that enable semantic units to move between these levels, and (iii) mapping relations that preserve semantic equivalence across alternative representations. Within this architecture, semantic units located at different levels may represent the same underlying proposition while differing in their degree of formal structure and machine-interpretability.

In this sense, the Semantic Ladder operationalizes the representation-formalization dimension of the Semantic Units Framework by introducing an explicit ordering and transformation logic over alternative representations. It thereby establishes progressive semantic formalization as a guiding design principle for data and knowledge infrastructures, enabling knowledge to be captured in flexible, human-accessible forms and incrementally enriched into fully formal, machine-interpretable representations without loss of meaning or provenance.

## 4.2 Levels of the Semantic Ladder

The Semantic Ladder organizes semantic units according to increasing degrees of semantic formalization. Let

$$L = \{L_1, L_2, L_3, L_4, L_5\}$$

denote the ordered set of **representation levels**, where increasing indices correspond to increasing levels of semantic explicitness and machine-interpretability. Each level represents a distinct form of data and knowledge representation, while semantic units located at different levels may still represent the same underlying proposition. The overall structure of the Semantic Ladder is illustrated in Figure 6.

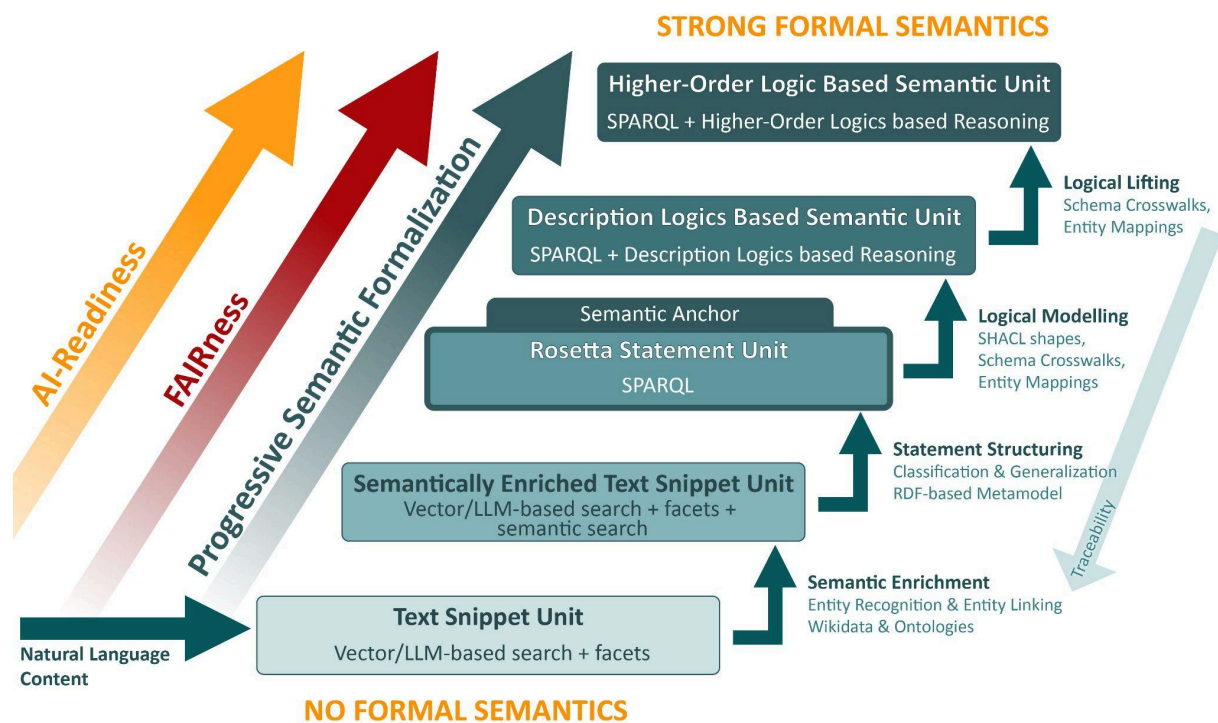

**Figure 6: The Semantic Ladder for progressive semantic formalization of data and knowledge**: The Semantic Ladder organizes semantic units according to increasing levels of semantic formalization while preserving semantic continuity between representations of the same semantic content. At the lowest level, **Text Snippet Units** capture natural language statements without formal semantics. Through semantic enrichment (e.g., entity recognition and entity linking to resources such as Wikidata or domain ontologies), text snippets become **Semantically Enriched Text Snippet Units**, enabling improved search and semantic annotation. Further structuring through statement structuring transforms enriched text into

**Rosetta Statement Units**, which represent canonical statement structures modelled in RDF that function as semantic anchors linking linguistic (lower levels) and logical representations (higher levels). Through logical modelling, RDF-based Rosetta Statement units can be mapped to **Description Logics–based Semantic Units**, typically implemented using ontology frameworks such as OWL and queried using SPARQL. Finally, logical lifting extends these representations into **Higher-Logic Semantic Units**, enabling more expressive reasoning using rule-based or higher-order logical frameworks. Across the ladder, semantic units remain interpretable to humans through dynamic textual and graphical representations while becoming increasingly findable, interoperable, and in general machine-actionable. The progression supports incremental knowledge formalization, improving interoperability, reasoning capability, **FAIRness**, and **AI-Readiness**, while maintaining alignment with the CLEAR principle.

## $L_1$ – Text Snippet Units

At the lowest level of formalization, semantic units may consist of text snippets extracted from natural language sources. These units represent statements or knowledge claims in their original linguistic form without additional semantic enrichment. Examples include fragments from scientific publications, field notes, or textual annotations. Text snippet units preserve the original wording of knowledge claims and allow them to be referenced, cited, and contextualized within the data and knowledge space while maintaining their full linguistic expressiveness.

## $L_2$ – Semantically Enriched Text Snippet Units

The second level introduces text snippet units that are augmented with lightweight semantic annotations. Such annotations may include recognized entities, references to ontology terms or controlled vocabularies, and additional semantic metadata. This level introduces a limited degree of semantic structure while preserving the original textual formulation of the statement. Even modest levels of semantic annotation can significantly improve discoverability and interoperability of textual knowledge, reflecting the insight attributed to James Hendler that "*a little semantics goes a long way.*"

## $L_3$ – Rosetta Statement Units

The third level defines structured representations of statements using the Rosetta Statement model. Rosetta Statements decompose propositions into canonical semantic roles that represent the functional components of a statement. By explicitly representing the internal structure of statements, Rosetta Statement units provide a technology-independent representation of statement semantics. Within the Semantic Ladder, Rosetta Statements function as semantic anchors that bridge natural language expressions and formal logical models. Because they capture the canonical structure of recurring statement types, they enable semantically equivalent data and knowledge to be aligned across different representation frameworks.

## $L_4$ – Description Logics-based Semantic Units

At the level, semantic units are represented using Description Logics-based frameworks, typically implemented using ontology languages such as OWL in RDF-based knowledge graphs. These representations embed the semantic components of statements within formally defined ontological relations. As a result, the knowledge graph becomes amenable to automated reasoning, enabling computational systems to perform logical inference, detect inconsistencies, and derive implicit knowledge from explicitly represented statements.

## L₅ – Higher-Logic Semantic Units

The highest level of the Semantic Ladder consists of semantic units represented using logical frameworks that extend beyond Description Logics. These may include rule-based reasoning systems, logic programming approaches, or other higher-order logical representations. Such frameworks allow more expressive reasoning over data and knowledge but typically require greater modelling effort and more complex semantic structures.

Together, these levels define a progression from minimally structured natural language representations to highly formal logical models, enabling the same semantic content to be expressed within increasing semantic explicitness and machine-interpretability.

## 4.3 Rosetta Statements as Semantic Anchors

Natural language and formal logical models represent complementary but fundamentally different modes of knowledge representation. Bridging these two representational domains is a central challenge for semantic data and knowledge infrastructures. Rosetta Statement address this challenge by providing canonical representations of recurring statement structures. Each Rosetta Statement schema captures the semantic roles that constitute a particular type of proposition, thereby making the internal structure of statements explicit. Within the Semantic Ladder, Rosetta Statements thus constitute the central bridging layer connecting linguistic representations and formal logical models, thereby enabling consistent alignment of semantically equivalent content across representation levels.

In this role, Rosetta Statement units act as semantic anchors that connect alternative representations of the same content. Natural language expressions can be mapped to Rosetta Statement structures through statement structuring, while ontology-based representations can be derived from the same structures through logical modelling. By linking both linguistic and formal representations to a shared intermediate structure, Rosetta Statements establish a stable reference point that preserves the continuity of meaning across the Semantic Ladder.

Within the overall architecture, Rosetta Statement units support bidirectional integration. They enable natural language content to be progressively formalized into machine-interpretable representations while ensuring that formal models remain grounded in human-interpretable statement structures. In this way, they provide the structural and semantic continuity required for progressive semantic formalization across the Semantic Ladder.

## 4.4 Ladder Transformations

The Semantic Ladder not only defines levels of semantic formalization but also specifies the transformations through which semantic content moves between them (Fig. 6). These transformations progressively increase semantic structure and computational interpretability while preserving underlying meaning.

Four principal transformation types can be distinguished: **semantic enrichment**, **statement structuring**, **logical modelling**, and **logical lifting**. Each corresponds to a transition between adjacent ladder levels and introduces additional semantic explicitness into the representation of data and knowledge. Several of these transformations can be partially automated using natural language processing, information extraction, and ontology mapping techniques.

### Semantic Enrichment ($L_1 \to L_2$)

The first transformation introduces lightweight semantic annotations to natural language text snippets. This may include entity recognition, linking textual expressions to ontology terms or controlled vocabulary identifiers, and the addition of semantic metadata. As a result, textual content becomes partially structured while preserving its original linguistic form, improving findability and accessibility.

### Statement Structuring ($L_2 \to L_3$)

The second transformation converts semantically enriched text into structured Rosetta Statement representations. In this step, the semantic roles of statement components are identified and mapped to positions within a Rosetta Statement structure, making the internal organization of propositions explicit. These representations can be implemented in RDF, with corresponding schemata defined as SHACL shapes using ontology terms. This enables semantic interoperability across instances of Rosetta Statement units of the same class and supports querying via SPARQL, substantially enhancing interoperability and reuse.

### Logical Modelling ($L_3 \to L_4$)

The third transformation maps Rosetta Statement units to ontology-based semantic graphs expressed in Description Logics frameworks such as OWL. Here, semantic components of statements are connected through formally defined relations, enabling automated reasoning, consistency checking, and logical inference.

### Logical Lifting ($L_4 \to L_5$)

The final transformation extends Description Logics-based representations into more expressive logical frameworks, including rule-based reasoning systems and logic programming approaches. These enable more advanced forms of inference beyond the expressive scope of standard Description Logics.

Taken together, these transformations enable semantic content to be moved progressively from natural language representations toward increasingly formal logical models. Importantly, the Semantic Ladder does not require immediate formalization at higher levels of formalization. Instead, data and knowledge can be incrementally enriched as additional semantic structure becomes available, while preserving semantic continuity and maintaining traceable provenance across representation levels.

## 4.5 Schema Crosswalks and Entity Mappings

As semantic units move between levels of the Semantic Ladder, preserving semantic equivalence requires explicit mappings between representation structures and the terminologies they employ. These mappings ensure that transformations across levels maintain both the propositional structure and the meaning of the represented content. Two complementary mapping types support this process: schema crosswalks and entity mappings.

**Schema crosswalks** define correspondences between alternative representation schemata that describe the same type of information. For example, a Rosetta Statement schema expressed as a SHACL shape can be mapped to a corresponding OWL-based schema, specifying how semantic roles and the statement positions align with the classes, properties, and constraints of the target representation. Such mappings enable structured statements to be translated into formal logical representations while preserving their underlying semantic structure, supporting propositional interoperability (3).

**Entity mappings**, in contrast, align the terminologies used within these representations. Different knowledge infrastructures often rely on distinct ontologies, controlled vocabularies, or knowledge graph properties to describe similar concepts. Entity mappings establish equivalences or hierarchical relations between such terms, allowing semantically equivalent statements to be recognized even when different identifiers or terminologies are used, supporting terminological interoperability (3).

Together, schema crosswalks and entity mappings ensure that semantic units remain aligned across representation levels despite differences in modelling approaches and terminologies. By explicitly linking both structural schemata and domain vocabularies, they enable multiple representations of the same content while maintaining semantic continuity across the Semantic Ladder.

## 4.6 Semantic Continuity and Traceability

Semantic units located at different levels of the Semantic Ladder may represent **semantically equivalent content** while differing in their degree of formalization, machine-interpretability, and machine-actionability. Within the Semantic Units Framework, these alternative representations remain explicitly linked through shared identifiers and mapping relations, allowing them to coexist within a unified semantic knowledge space.

This architecture establishes **semantic continuity** across representation levels. Natural language expressions, structured statements, and formal logical models can be treated as equivalent representations that preserve meaning while providing increasing levels of semantic explicitness and machine support.

A key consequence of this design is **bidirectional traceability**. Formal representations at higher levels can be traced back to their originating Rosetta Statement units and ultimately to natural language expressions, while lower-level representations can be progressively enriched and formalized. This ensures that machine-interpretable models remain grounded in **human-interpretable statements**, supporting transparency, explainability, and reproducibility of knowledge modelling processes, resulting in **CLEAR data and knowledge spaces**.

The Semantic Ladder also frames **FAIRness as a continuum** rather than a binary Boolean property. Semantic units at different levels exhibit varying degrees of findability, accessibility, interoperability, and reusability, with higher levels enabling greater machine-actionability. Ladder transformations provide a structured pathway for progressively increasing FAIRness without requiring full formalization at the point of data entry. In this sense, the Semantic Ladder provides a structural mechanism for organizing the FAIR continuum into **semi-permeable levels of semantic**

**formalization**, across which data and knowledge can progressively move through ladder transformations.

This progression is particularly relevant for collaborative and community-driven knowledge production. Contributors can engage with a shared data and knowledge space at different levels of formalization, allowing content to be captured in accessible forms and incrementally enriched over time. In this way, the Semantic Ladder supports both broad participation and the gradual development of semantically rich, machine-interpretable knowledge.

By organizing semantic units into progressively formal representation levels while preserving continuity and traceability, the Semantic Ladder provides a unifying architecture that connects human communication with machine-interpretable data and knowledge infrastructures.

# 5 Evolution of Semantic Data and Knowledge Spaces

## 5.1 Incremental Knowledge Graph Construction

Knowledge graphs are rarely constructed as fully specified systems from the outset. In practice, data and knowledge emerge incrementally, often originating as unstructured or semi-structured content that only gradually acquires formal semantic structure. Traditional ontology-driven approaches, which require data to conform to predefined schemata and formal representations at the point of entry, impose substantial modelling overhead and create barriers to participation (*semantic parsing burden*), particularly in open, collaborative, and cross-domain settings (*dynamic knowledge graph construction challenge* (7)).

The Semantic Units Framework, combined with the Semantic Ladder, enables an alternative mode of knowledge graph construction based on **incremental formalization**. Instead of requiring immediate transformation into fully formalized representations, semantic content be introduced in its original form and progressively refined over time. This allows knowledge graphs to develop through continuous extension and restructuring rather than through upfront design.

In this process, semantic units may initially capture knowledge in minimally structured forms and are subsequently enriched, structured, and formalized as modelling needs evolve. These transformations do not replace earlier representations but extend them, resulting in multiple coexisting representations of the same content at different levels of formalization. Because these representations remain explicitly linked, the knowledge graph preserves semantic continuity while increasing its level of formal structure.

This incremental construction paradigm fundamentally shifts how knowledge graphs are developed. Rather than being engineered as static systems defined by complete ontological models, they can be **grown dynamically**, accommodating new data, contributors, and modelling perspectives without requiring immediate standardization. As a result, the Semantic Ladder supports scalable knowledge graph construction in heterogeneous and evolving environments, where semantic structure emerges progressively through use, while maintaining interoperability and traceability.

## 5.2 Purpose-Driven Semantic Enrichment

A key implication of the Semantic Ladder is that semantic enrichment is **not a linear process aimed at maximal formalization**, but a **purpose-driven activity** guided by the requirements of specific analytical tasks. Different uses of data and knowledge require different degrees of semantic structure, and in many cases, lightweight semantic representations are sufficient to support meaningful interaction.

In contrast to traditional ontology-driven approaches, which often treat formal semantic modelling as a prerequisite for integration and analysis, the Semantic Ladder allows representations to remain at the level of formalization that is appropriate for their intended use. As a result, semantic enrichment can be applied selectively, focusing effort where additional structure provides clear benefits.

This relationship between analytical tasks and levels of formalization can be summarized as follows (Table 1). Basic information retrieval can be supported by minimally structured representations, while semantic search benefits from lightweight annotation. More advanced operations, such as structured querying and data integration, require explicitly structured representations, and logical reasoning depends on fully formalized semantic models.

Table 1: Typical analytical tasks supported at different levels of semantic formalization in the Semantic Ladder.

| Use case | Typical ladder level |
| --- | --- |
| Text retrieval | $L_1$–$L_2$ |
| Semantic search | $L_2$–$L_3$ |
| Structured querying and integration | $L_3$–$L_4$ |
| Logical reasoning and advanced inference | $L_4$–$L_5$ |

This perspective reframes semantic enrichment as a **context-dependent optimization problem** rather than a uniform modelling requirement. Instead of striving for complete formalization across all data, knowledge infrastructures can allocate modelling effort strategically, enabling efficient scaling while preserving the ability to support advanced analytical capabilities where needed, and still allowing less formal representations to remain integrated within the same semantic knowledge space.

By allowing multiple levels of semantic expressivity to coexist, the Semantic Ladder supports knowledge spaces in which data can be incrementally enriched in response to evolving use cases, rather than being constrained by upfront modelling requirements.

## 5.3 Collaborative Knowledge Development

The Semantic Ladder also enables a model of **collaborative knowledge development** in which contributors with different expertise levels can participate in the construction and refinement of semantic knowledge spaces. In many real-world settings, knowledge is produced by heterogeneous communities, including domain experts, data curators, and technical specialists, each of whom engages with data at different levels of abstraction and formalization.

By allowing semantic content to enter the system at multiple levels of formalization, the Semantic Ladder supports a **distributed curation process**. Contributors are not required to produce fully formalized representations at the point of entry. Instead, they can contribute at the level that matches their expertise and available resources. For example, domain experts may provide natural language descriptions or observational data, while curators and knowledge engineers may subsequently introduce semantic annotations, structured representations, or formal logical models.

This results in a division of labour across representation levels. Lower levels of the Semantic Ladder ($L_1$–$L_2$) primarily support knowledge capture and initial annotation, while higher levels ($L_3$–$L_5$) support structuring, integration, and formal reasoning. Because these representations remain explicitly linked, contributions made at different stages can be integrated into a coherent and evolving knowledge space.

The layered structure of the Semantic Ladder further enables the integration of automated and AI-assisted processes into collaborative workflows. Tools for entity recognition, linking, and statement extraction can support early-stage enrichment, while more advanced modelling and validation processes can operate on increasingly formal representations. In this way, human and machine contributions can be combined across different levels of semantic formalization.

This collaborative model allows knowledge infrastructures to scale through incremental and distributed contributions, rather than relying on centralized modelling efforts. Such models are particularly relevant for large-scale knowledge infrastructures that depend on community participation, including citizen science initiatives and community-driven knowledge bases. By supporting participation at multiple levels of formalization, the Semantic Ladder enables knowledge spaces to grow organically while maintaining semantic coherence and enabling progressive refinement over time.

# 6 Supporting Hybrid AI–Knowledge Systems

## 6.1 Vector Representations

Vector-based representations have become a central component of modern AI systems, enabling similarity-based retrieval and clustering of textual and structured content in high-dimensional embedding spaces. Rather than relying on explicitly defined semantic relations, these approaches capture contextual similarity by encoding content as numerical vectors, allowing systems to identify related concepts or statements based on proximity in embedding space.

Within the Semantic Units Framework, vector representations can be associated with semantic units as a **complementary representation layer**. Independent of the representation level, each semantic unit has a natural language representation of its content (either directly or via their dynamic label) that can be embedded using language models, enabling similarity-based search, grouping, and exploration across large collections of units. These capabilities are particularly valuable at lower levels of formalization, where semantic structure is only partially specified.

Importantly, vector representations do not replace the explicit semantic structures provided by higher levels ($L_3$–$L_5$). In other words, vector representations are **not part of the Semantic Ladder itself**. They do not constitute an additional level of semantic formalization, but instead operate in parallel to the ladder as an alternative mode of representation. While ladder levels ($L_1$–$L_5$) describe increasing degrees of explicit semantic structure, vector embeddings capture implicit similarity relationships that are not directly encoded in symbolic form.

As a result, vector-based and symbolic representations provide complementary capabilities. Embeddings support flexible retrieval and discovery based on contextual similarity, whereas structured semantic representations enable precise querying, interoperability, and logical reasoning. By associating vector embeddings with semantic units across different ladder levels, knowledge infrastructures can combine these capabilities within a unified system.

This integration establishes the basis for hybrid AI–knowledge systems in which similarity-based and logic-based methods operate on shared semantic content, enabling both exploratory and formally grounded interactions with data and knowledge.

## 6.2 LLM-Assisted Semantic Enrichment

Large language models (LLMs) provide powerful mechanisms for transforming natural language content into increasingly structured representations, making them well suited to support the progression of semantic units along the Semantic Ladder. Rather than introducing a new representation paradigm, LLMs act as **operational tools that facilitate ladder transformations**, particularly at lower and intermediate levels of formalization.

At the transition from $\mathbf{L_1}$ to $\mathbf{L_2}$, LLMs can support **semantic enrichment** by identifying entities, concepts, and relationships within text and linking them to controlled vocabularies or knowledge graph identifiers. This process introduces lightweight semantic structure while preserving the original linguistic formulation of the content. At the transition from $\mathbf{L_2}$ to $\mathbf{L_3}$, LLMs can assist in **statement structuring** by identifying syntactic positions of a proposition, extracting their semantic roles, and mapping them to canonical representations corresponding to Rosetta Statement units.

More generally, LLMs can contribute to multiple stages of the transformation pipeline by generating candidate annotations, structured statements, and, in some cases, preliminary logical representations. These outputs, however, require validation and refinement, as the correctness and consistency of formal semantic representations cannot be guaranteed through automated generation alone.

The Semantic Ladder provides the architectural context in which these capabilities can be systematically integrated. By aligning LLM-assisted processes with specific transformation steps, knowledge infrastructures can incorporate AI-driven enrichment while maintaining explicit control over representation quality and semantic consistency. In this setting, LLMs function as **scalable transformation mechanisms**, while human experts ensure the reliability and interpretability of the resulting semantic structures.

This combination of automated extraction and human validation enables the efficient expansion of semantic knowledge spaces, allowing large volumes of natural language content to be progressively incorporated into structured and interoperable representations.

## 6.3 Hybrid Symbolic–Vector Data and Knowledge Systems

Modern AI systems increasingly rely on the combination of multiple representation paradigms to balance complementary strengths. Symbolic representations, vector-based embeddings, and natural language each provide distinct capabilities for representing and processing data and knowledge, but no single paradigm is sufficient on its own. Integrating these approaches within a coherent architecture therefore represents a central challenge for next-generation knowledge systems.

Symbolic representations, such as ontology-based knowledge graphs and structured semantic statements, provide explicit, formally interpretable models that support precise querying, semantic interoperability, and logical reasoning. Vector-based representations, in contrast, encode semantic similarity in high-dimensional embedding spaces, enabling flexible retrieval, clustering, and similarity-based exploration. Natural language representations remain essential for human communication and interpretation, serving as the primary interface through which data and knowledge is created and consumed. These representational paradigms therefore offer complementary strengths (Table 2).

Table 2: Complementary strengths of different knowledge representation paradigms in hybrid AI-knowledge systems.

| Representation paradigm | Primary strength |
| --- | --- |
| Symbolic representations | Logical reasoning and formal interoperability |
| Vector representations | Similarity-based retrieval and semantic proximity |
| Natural language representations | Human-interpretability and communication |

The Semantic Ladder provides a **structured integration mechanism** for these paradigms by organizing semantic units along levels of increasing formalization. Lower ladder levels ($L_1$–$L_2$) naturally accommodate natural language representations and their associated vector embeddings, supporting similarity-based retrieval and exploratory interaction. Higher levels ($L_3$–$L_5$) introduce progressively more explicit semantic structure through Rosetta Statement representations and ontology-based logical models, enabling formal reasoning and semantic integration.

In this architecture, vector-based and symbolic representations are not competing alternatives but operate on shared semantic units at different levels of formalization. Vector embeddings provide a complementary layer that enables similarity-based access to content across the ladder, while symbolic representations provide the explicit structure required for precise interpretation and reasoning and natural language representations for human-interpretability of the content. The Semantic Ladder thus establishes a **bridge between statistical and symbolic approaches**, linking similarity-based and logic-based methods within a unified knowledge space.

This integration enables hybrid systems in which natural language content, embedding-based representations, and formal semantic models can be combined within a single workflow. Knowledge can be explored through similarity-based retrieval, structured through intermediate representations, and formalized for reasoning and interoperability, all while remaining connected through shared semantic units.

By providing this unifying architecture, the Semantic Ladder offers a pathway for integrating modern machine learning approaches with symbolic knowledge representation, supporting data and knowledge infrastructures that combine scalability, flexibility, and formal semantic rigour.

# 7 Discussion

## 7.1 Addressing the Semantic Parsing Burden

A central challenge in the development of semantic data and knowledge infrastructures is the semantic parsing burden, that is, the substantial effort required to translate natural language content into formally structured representations. Traditional ontology-driven approaches address this challenge by enforcing explicit semantic modelling at the point of data entry. While this enables precise and interoperable knowledge graphs, it also creates a fundamental tension between semantic rigour and practical scalability.

The Semantic Ladder reframes this challenge by shifting semantic parsing from a **prerequisite** to a **progressive process**. Rather than requiring complete formalization upfront, it allows semantic structure to emerge incrementally as data and knowledge are enriched over time. This transformation changes the role of semantic modelling from a gatekeeping requirement to an ongoing process of refinement.

This perspective aligns with the widely cited observation that "a little semantics goes a long way," but extends it by embedding this principle into a formal architectural framework. Even minimal semantic annotations can provide immediate benefits for discovery and integration, while more expressive representations can be introduced selectively as needed. In this way, semantic expressivity becomes a **scalable resource** rather than a fixed requirement.

At the same time, this shift introduces new challenges. Progressive formalization requires mechanisms for maintaining semantic consistency across representations and for managing the coexistence of multiple levels of formalization within the same knowledge space. It also places greater emphasis on curation workflows and validation processes, particularly when automated methods are used to generate intermediate representations.

Overall, the Semantic Ladder transforms the semantic parsing burden from a bottleneck into a **manageable and distributed process**, enabling knowledge infrastructures to balance formal rigour with practical scalability. By decoupling knowledge capture from full formalization, it opens new pathways for participation while preserving the long-term goal of semantically rich and interoperable data and knowledge systems.

## 7.2 Term and Schema Alignment for Semantic Interoperability

Achieving semantic interoperability across heterogeneous data and knowledge sources remains a persistent challenge for large-scale information infrastructures. Different communities often describe similar phenomena using distinct ontologies, vocabularies, and modelling conventions, reflecting varying priorities between formal rigour, usability, and scalability. As a result, integrating such sources typically requires extensive alignment efforts at both the terminological and schema levels.

Traditional approaches to interoperability rely on establishing mappings between pairs of schemata or pairs of terms. However, as the number of data sources increases, this strategy becomes increasingly difficult to maintain, leading to a rapid growth in mapping complexity. This challenge is particularly pronounced in open and collaborative environments, where modelling practices evolve dynamically and cannot be centrally controlled.

The Semantic Units Framework introduces an alternative perspective by shifting the focus from pairwise alignment to **alignment through shared semantic structures**. By representing recurring statement types in a canonical form, Rosetta Statement units provide common reference points to which heterogeneous representations can be related. Instead of directly mapping all schemata to one another, different modelling approaches can be aligned through these shared structures, reducing the overall complexity of interoperability (from $n^2$ to $2n$).

This shift has important implications. It allows heterogeneous modelling paradigms, including highly formal ontology-based systems and more flexible, community-driven knowledge graphs, to coexist within a unified semantic space without requiring strict standardization. Interoperability becomes a matter of **anchoring diverse representations to common semantic patterns**, rather than enforcing uniform modelling practices across all contributors.

At the same time, this approach does not eliminate the need for mappings, but redistributes the effort in a more scalable way. It emphasizes the importance of maintaining stable and well-defined reference structures while allowing variability at the level of individual representations. In this sense, the Semantic Ladder extends interoperability from a static alignment problem to a **dynamic and continuously evolving process**.

By enabling alignment through shared semantic anchors, the framework provides a mechanism for managing the growing diversity and scale of semantic data and knowledge ecosystems, while preserving both flexibility and formal interpretability.

## 7.3 Implications for FAIR, CLEAR, and AI-Ready Data and Knowledge Spaces

The Semantic Ladder also has important implications for how FAIR and AI-ready data and knowledge infrastructures can be developed in practice. While the FAIR Principles provide widely adopted guidelines for improving the findability, accessibility, interoperability, and reusability of digital resources, their implementation is often associated with substantial upfront semantic modelling effort.

The Semantic Ladder reframes this challenge by embedding FAIRness within a process of **progressive formalization**. Rather than treating FAIR compliance as a prerequisite, it allows data and knowledge to gradually acquire FAIR characteristics as they are enriched and structured over time. In this perspective, FAIRness emerges incrementally, with increasing levels of semantic explicitness enabling progressively more advanced forms of interoperability and reuse.

This progression is closely aligned with the requirements of AI-ready content. Lower levels of formalization support large-scale ingestion and similarity-based processing, while higher levels provide the explicit structure required for reasoning, validation, and integration across heterogeneous sources. The ladder therefore establishes a pathway through which data and knowledge can evolve from human-readable content to machine-actionable knowledge without requiring immediate formalization.

At the same time, this architecture highlights the importance of balancing machine-actionability with human-interpretability. While increasing formalization enhances computational usability, maintaining links to natural language representations ensures that knowledge remains accessible and interpretable to human users. In this respect, the Semantic Ladder complements the FAIR Principles with considerations aligned with the CLEAR Principle, supporting data and knowledge representations that are not only machine-actionable but also transparent and explainable.

By integrating progressive formalization with both machine-oriented and human-oriented requirements, the Semantic Ladder provides a framework in which FAIRness and AI-readiness become **emergent properties of evolving knowledge systems**, rather than fixed conditions imposed at the point of data creation, while human-interpretability and thus CLEARness is always guaranteed across all levels of representation.

# 8 Conclusion

The increasing volume and heterogeneity of digital data and knowledge continue to challenge the design of scalable semantic information infrastructures. While much of this information is created and communicated in natural language, machine-interpretable systems rely on formally structured representations to enable interoperability, reasoning, and reuse. Bridging this gap remains a central obstacle for the development of integrated data and knowledge ecosystems.

In this work, we introduced an architectural framework that addresses this challenge through progressive semantic formalization. Building on the Semantic Units Framework, the Semantic Ladder enables data and knowledge to evolve from informal textual representations toward increasingly formal semantic models while preserving continuity of meaning across representations. By organizing semantic units into levels of formalization and defining transformations between them, the framework replaces the requirement for upfront formalization with a model of incremental enrichment. This approach lowers barriers to participation, supports the dynamic growth of knowledge graphs, and maintains traceability between natural language sources and formal representations. At the same time, the combination of modular semantic units, canonical statement structures, and explicit mappings provides a basis for interoperability across heterogeneous vocabularies, schemata, and modelling paradigms.

Beyond its implications for semantic data modelling, the framework supports the integration of symbolic, vector-based, and natural language representations within hybrid AI–knowledge systems. By enabling these complementary paradigms to coexist within a unified semantic space, the Semantic Ladder establishes a foundation for information infrastructures that combine human interpretability, flexible similarity-based retrieval, and formal semantic reasoning.

Taken together, these contributions suggest a shift toward more adaptive and inclusive semantic infrastructures in which data and knowledge can be captured, enriched, and formalized progressively. Future work will be required to develop scalable implementations, automated enrichment workflows, and domain-specific applications of the framework. In this perspective, the Semantic Ladder provides a pathway toward data and knowledge systems in which natural language, semantic models, and AI methods interact productively within a shared and evolving semantic architecture.

# Acknowledgements

The author acknowledges the use of an AI-based language model to assist with language editing, phrasing, and manuscript structuring. The AI system was not used to generate scientific content, and all ideas, concepts, analyses, and conclusions presented in this work are those of the author.